\documentclass[conference,compsoc]{IEEEtran}
\IEEEoverridecommandlockouts
\usepackage{cite}
\usepackage{amsmath,amssymb,amsfonts}
\usepackage{algorithmic}
\usepackage{graphicx}
\usepackage{textcomp}
\usepackage{xcolor}
\usepackage{comment}
\def\BibTeX{{\rm B\kern-.05em{\sc i\kern-.025em b}\kern-.08em
    T\kern-.1667em\lower.7ex\hbox{E}\kern-.125emX}}

\author{\IEEEauthorblockN{1\textsuperscript{st} Gjorgjina Cenikj}
\IEEEauthorblockA{\textit{Computer Systems Department} \\
\textit{Jo\v{z}ef Stefan Institute}\\
\textit{Jo\v{z}ef Stefan International}\\ \textit{Postgraduate School}\\
Ljubljana, Slovenia \\
gjorgjina.cenikj@ijs.si}
\and
\IEEEauthorblockN{2\textsuperscript{nd} Gašper Petelin}
\IEEEauthorblockA{\textit{Computer Systems Department} \\
\textit{Jo\v{z}ef Stefan Institute}\\
\textit{Jo\v{z}ef Stefan International}\\ \textit{Postgraduate School}\\
Ljubljana, Slovenia \\
gasper.petelin@ijs.si}
\and
\IEEEauthorblockN{3\textsuperscript{rd} Tome Eftimov}
\IEEEauthorblockA{\textit{Computer Systems Department} \\
\textit{Jo\v{z}ef Stefan Institute}\\
Ljubljana, Slovenia \\
tome.eftimov@ijs.si}
}
\usepackage{subcaption}
\usepackage{geometry}
\usepackage{graphicx}
\usepackage{url}
\begin{document}
\title{ClustOpt: A Clustering-based Approach for Representing and Visualizing the Search Dynamics of Numerical Metaheuristic Optimization Algorithms}

  \IEEEpubidadjcol
\maketitle

\begin{abstract}
Understanding the behavior of numerical metaheuristic optimization algorithms is critical for advancing their development and application. Traditional visualization techniques, such as convergence plots, trajectory mapping, and fitness landscape analysis, often fall short in illustrating the structural dynamics of the search process, especially in high-dimensional or complex solution spaces. To address this, we propose a novel representation and visualization methodology that clusters solution candidates explored by the algorithm and tracks the evolution of cluster memberships across iterations, offering a dynamic and interpretable view of the search process. Additionally, we introduce two metrics -- algorithm stability and algorithm similarity -- to quantify the consistency of search trajectories across runs of an individual algorithm and the similarity between different algorithms, respectively. We apply this methodology on a set of ten numerical metaheuristic algorithms, revealing insights into their stability and comparative behaviors, thereby providing a deeper understanding of their search dynamics.
\end{abstract}

\begin{IEEEkeywords}
algorithm trajectory representations, optimization algorithm analysis
\end{IEEEkeywords}
\section{Introduction}

Visualization techniques are a critical means of shedding light on the behavior of metaheuristic numerical optimization algorithms. Conventional methods such as convergence analysis, trajectory visualizations, and fitness landscape analysis provide valuable insights into aspects like convergence speed, diversity, and solution quality. However, these approaches often fail to capture the structural dynamics of the search process, particularly in high-dimensional or complex spaces. Existing methods rarely address the location of the solution candidates in the search space, which can reveal crucial information about the exploratory and exploitative strategies of an algorithm.

We propose ClustOpt, a novel representation and visualization methodology for metaheuristic numerical population-based optimization algorithms, that focuses on clustering solution candidates explored by optimization algorithms. The methodology groups these candidates based on their similarity in the solution space and visualizes the evolution of cluster memberships across iterations. It provides a dynamic and interpretable representation of the search process, highlighting patterns such as the emergence, dominance, or disappearance of clusters over time. The resulting visualizations offer insights into the algorithms' navigation of the search space, enabling a deeper understanding of their behavior.
Unlike traditional visualization techniques that focus solely on visualizations, our approach transforms visualizations into structured representations that allow for efficient comparison across multiple algorithms. It enables large-scale benchmarking of optimization algorithms' sensitivity to initialization and the analysis of their similarity based on detailed search trajectories.
We demonstrate the methodology's applicability for analyzing the behaviour of 10 single-objective continuous optimization algorithms on the Black Box Optimization Benchmarking (BBOB)~\cite{bbob} suite.

The remainder of the paper is organized as follows. In Section~\ref{sec:related_work}, we present the related work, while in Section~\ref{sec:methodology} we explain the methodology of constructing the representations. In Section~\ref{sec:results} we demonstrate how the representations can be visualized and used for different tasks. Finally, Section~\ref{sec:conclusion} concludes the paper and specifies directions for future work.

\textbf{Reproducibility:} The code is available at \url{https://github.com/gjorgjinac/ClustOpt}.


\section{Related work}
\label{sec:related_work}
A prevalent method for analyzing the behaviour of single-objective continuous optimization algorithms is convergence analysis~\cite{convergence_analysis_1, convergence_analysis_2}, which seeks to determine whether an algorithm converges to the global optimum and the speed at which it achieves convergence. However, this approach focuses on the quality of the solutions, ignoring their location in the search space.

One of the biggest challenges of visualizing the solutions explored by an algorithm during the search procedure is the large number of solutions being explored, especially when the problem is high dimensional. Consequently, several works have already attempted to reduce the high dimensional data into two dimensions using different dimensionality reduction techniques such as Principal Component Analysis~\cite{tea_visualizing, collins2003applying}, t-Distributed Stochastic Neighbor Embedding (t-SNE)~\cite{tea_visualizing, maaten2008visualizing}, Uniform Manifold Approximation and Projection for Dimension Reduction (UMAP)~\cite{tea_visualizing}, Sammon mappings~\cite{sammon1969nonlinear, pohlheim2006multidimensional}, and Self-Organizing Maps~\cite{amor2005intelligent}. An analysis of the ability of different dimensionality reduction techniques to preserve meaningful information about the population
dynamics can be found in~\cite{tea_visualizing}. While such approaches are useful for visualization purposes, they do not allow an automatic comparison of algorithm behavior and may be too overwhelming for manual examination when numerous algorithms, problems, and executions are compared.

Apart from methodologies based on dimensionality reduction, graph visualization approaches have also been proposed. For example, Local Optima Networks (LONs)~\cite{ochoa2014local} represent fitness landscapes in terms of a graph where local optima are nodes and search transitions are edges defined by an exploration search operator. Similarly, Search Trajectory Networks (STNs)~\cite{ochoa2020search}, are graph-based visualizations where nodes represent locations of the search space, not limited to local optima, while the edges signify the progression between these locations. In this case, the search space is partitioned into a predefined set of locations and an algorithm's optimization process is visualized by taking the locations of representative solutions from each iteration and connecting them into a graph structure. While our approach bares similarities to the STNs, the main difference is that we target the representation of the entire search trajectory (all candidate solutions within each population), while the STNs are representing only a single representative solution from each population (usually the best solution from each population). Additionally, STNs produce a graph-based visualization, while our approach produces a numerical representation of the trajectory which allows it to be visualized and compared to other trajectories using machine learning (ML) approaches.

\section{Methodology}
\label{sec:methodology}
We propose a methodology for visualizing algorithm trajectories and representing them in terms of numerical features. 
Given a fixed optimization problem $p$ of dimension $d$, and a set population-based algorithms $A = \{a_1, a_2, \dots \}$, executed multiple times with random seeds $R = \{r_1, r_2, \dots \}$ for a budget of $b$ iterations with a population size $s$ on the same problem, our proposed approach enables the analysis and comparison of the algorithm search behaviour.
The proposed approach can be used to analyze a single algorithm trajectory($|A|=1$, $|R|=1$), or multiple algorithm trajectories - different executions of the same algorithm ($|A|=1$, $|R|>1$) or executions of different algorithms ($|A|>1$, $|R| \geq 1$). 

It consists of four main steps:
\begin{itemize}
    \item Merging trajectories - The candidate solutions explored by all algorithms in all iterations of all executions are merged into a set $S$ containing $|A| \times |R| \times b \times s$ solution vectors of dimension $d$.
    \item Scaling - To account for the fact that the $d$ different dimensions of the $\mathbf{x} \in S$ candidate solutions may have different ranges of values, we first scale all candidate solutions to be in the range [0,1]. The scaling is done on the merged set of trajectories jointly (i.e., including the trajectories for all algorithms on all problem with different seeds) in order to allow a comparison of the results for different trajectories.
    \item Clustering - The scaled candidate solutions are then clustered into $c$ clusters. The search space is thereby partitioned into areas of interest which are shared across all executions of the algorithms.
    In our experiments, we determine the number of clusters using the elbow method and apply k-means clustering with Euclidean distance. We selected k-means clustering because it produces more spherical clusters, which are generally easier to interpret and analyze. We use a k-means++ initialization, which has been shown to speed up convergence and often improve the quality of clusters.
    \item Representation calculation - 
    The proposed representation is based on the number of solutions from each iteration of the trajectory which belong to each cluster. For this purpose, for each algorithm, execution, and iteration of the search process, we count the number of candidate solutions from the population generated by the algorithm in the particular iteration which belong to each of the $c$ clusters. The representation of a single trajectory is therefore a vector of $b \times c$ values, indicating the number of candidate solutions from each of the $b$ iterations which are placed into each of the $c$ clusters.


\end{itemize}

\section{Results}
\label{sec:results}
We demonstrate the applicability of the proposed methodology for the following purposes:
\textit{i)} Visualizing algorithm trajectories;
\textit{ii)} Analyzing the sensitivity of the optimization algorithm to its initialization;
\textit{iii)} Analyzing the similarity of different algorithms.

\subsection{Data}
We demonstrate the proposed methodology by analyzing the trajectories of 10 algorithms from the MEALPY~\cite{mealpy} Python library on the Black Box Optimization Benchmarking (BBOB) benchmark~\cite{bbob}. We use the first five instances from each of the 24 BBOB problems in dimensions $d = \{ 2, 5, 10 \}$. The instances from the same problem class differ in shifting, scaling, and/or rotation. The algorithms are executed with their default parameters, a population size of 50 and a budget of 10$d$ iterations (500$d$ function evaluations).

In particular, we select the following algorithms for analysis: BaseDE (Differential Evolution)~\cite{basede}, SADE (Self-Adaptive Differential Evolution)~\cite{sade}, JADE (Adaptive Differential Evolution With Optional External Archive)~\cite{jade}, SHADE (Success-history based parameter adaptation for Differential Evolution)~\cite{shade}, EnhancedAEO (Enhanced Artificial Ecosystem-based Optimization)~\cite{enhancedaeo}, ModifiedAEO (Modified Artificial Ecosystem-based Optimization)~\cite{modifiedaeo}, OriginalAEO (Artificial Ecosystem-based Optimization)~\cite{aeo}, AugmentedAEO (Augmented Ecosystem-based Optimization)~\cite{augmentedaeo}, HI\_WOA (Hybrid Improved Whale Optimization Algorithm)~\cite{hiwoa}, and OriginalWOA (Whale Optimization Algorithm)~\cite{woa}. 
We select these particular algorithms because they are variants of the Differential Evolution, Artificial Ecosystem-based Optimization, and the Whale Optimization Algorithm. We therefore expect some of the variants of the same algorithm to have similar behaviour, and we use this to confirm the validity of our representations. However, our representations are applicable to any population-based continuous optimization algorithm.

\subsection{Visualizing algorithm trajectories}

\begin{figure}
    \centering
    \includegraphics[width=\linewidth]{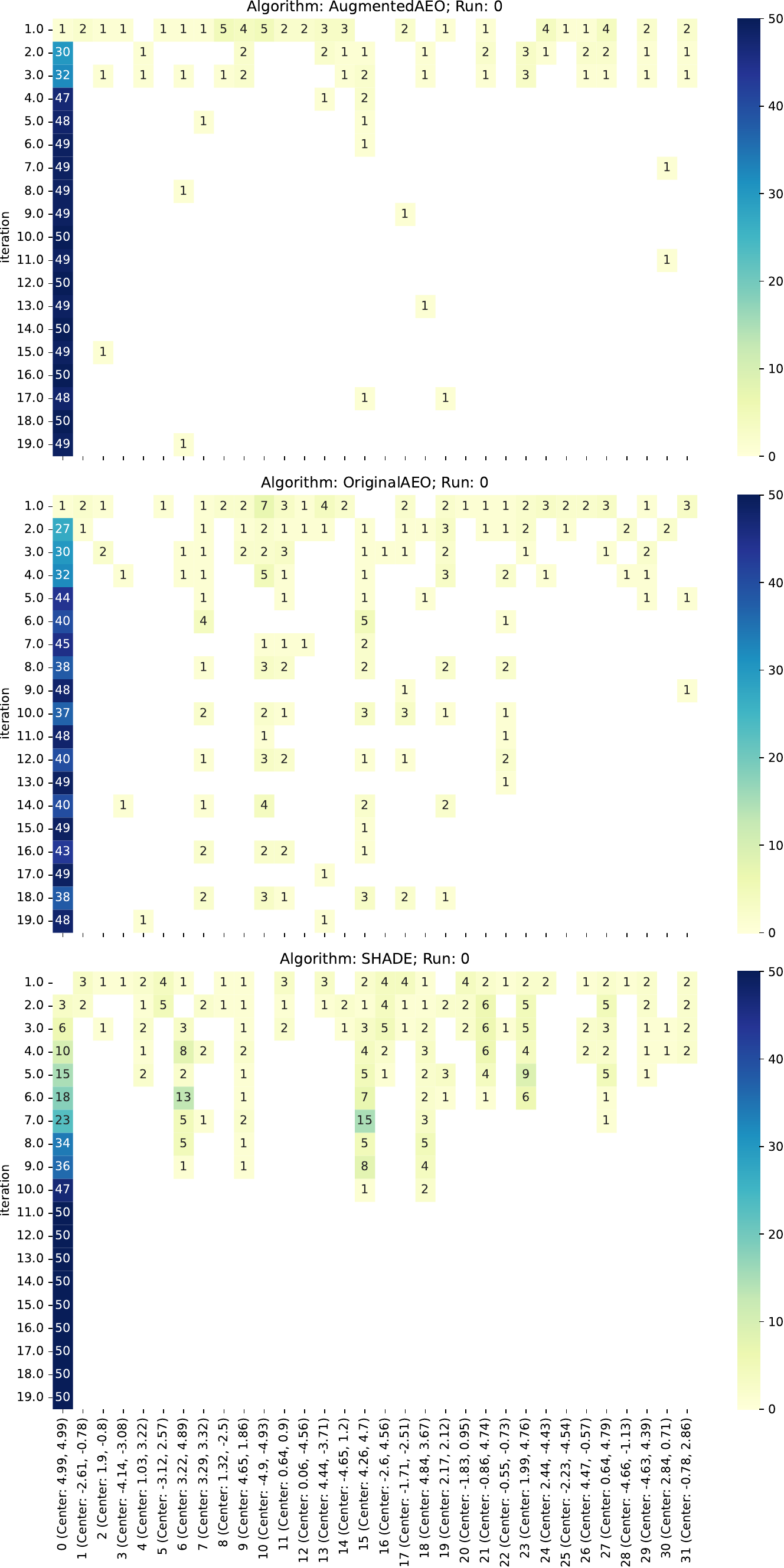}
    \caption{Visualizations of the trajectories of the OriginalAEO, AugmentedAEO and SHADE algorithms on the first instance of the fifth 2$d$ BBOB problem (Linear Slope)}
    \label{fig:single_traj_f5}
\end{figure}

\begin{figure}
    \centering
    \includegraphics[width=\linewidth]{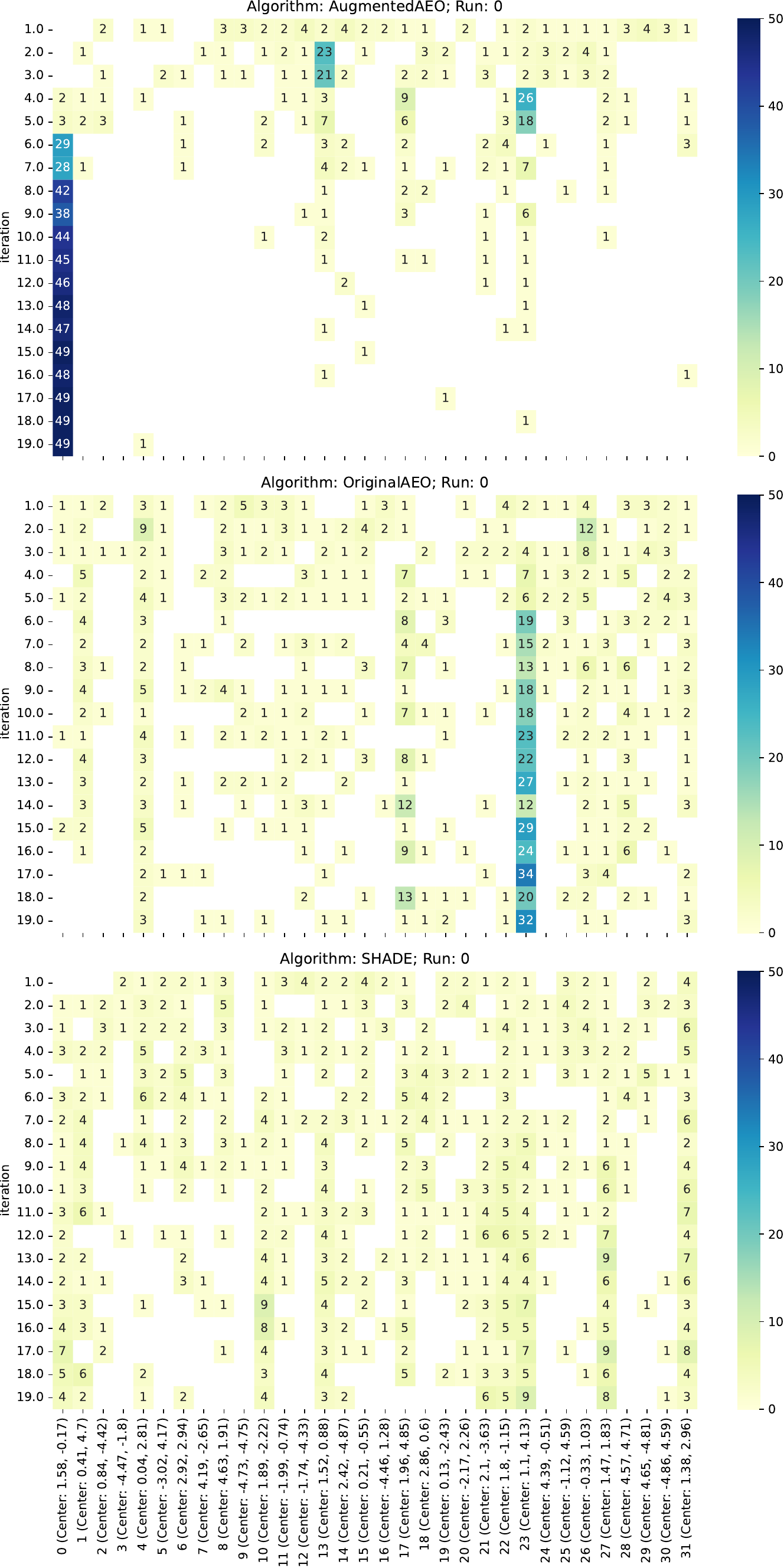}
    \caption{Visualizations of the trajectories of the OriginalAEO, AugmentedAEO and SHADE algorithms on the first instance of the 16th 2$d$ BBOB problem (Weierstrass)}
     \label{fig:single_traj_f16}
\end{figure}

In this section, we demonstrate how the proposed methodology can be used for visualizing a single algorithm trajectory and comparing it to trajectories of other algorithms. We start off with an example of how the proposed representations can be visualized. We demonstrate the example on 2$d$ problems, since it is straightforward to visualize such problems without a loss of information. However, the methodology is applicable to problems of higher dimensions, and in the following subsections we include analysis on 5$d$ and 10$d$ problems.
Figure~\ref{fig:single_traj_f5} depicts the representation of the OriginalAEO, AugmentedAEO, and SHADE algorithms on the first instance of the 5th 2$d$ BBOB problem (Linear Slope), while Figure~\ref{fig:single_traj_f16} shows their behaviour on the first instance of the 16th 2$d$ BBOB problem (Weierstrass). 

Each subplot refers to a different algorithm. The horizontal axis shows the cluster number, as well as the location of the cluster centroid. The vertical axis depicts the iteration number. The values in the heatmap cells denote the number of candidate solutions which were located in a particular cluster in a particular iteration. White cells denote empty clusters. The maximal value in the heatmap can be 50, i.e., the population size. 

Comparing the visualizations of the three algorithms on the 5th problem in Figure~\ref{fig:single_traj_f5}, we can see that in the first iterations, they are all exploring different clusters, however, they all converge to cluster zero with centroid (4.99,4.99). This is expected since the problem visualized is the linear slope function, which is convex and very simple to solve. 
On the other hand, looking at the algorithms' behaviour on the Weierstrass problem in Figure~\ref{fig:single_traj_f16}, we can see that the three algorithms have completely different outcomes. In the last iteration, the OriginalAEO algorithm has the majority of the candidate solutions in cluster 23, but is still doing some exploration of the other clusters. The AugmentedAEO algorithm ends up converging in cluster zero, with only a single candidate solution in cluster three. Finally, the SHADE algorithm is exploring multiple clusters and not focusing on a particular one. This behaviour is due to the fact that the Weierstrass function is highly rugged and moderately repetitive, and its global optimum is not unique~\cite{bbob}.

\begin{figure}
    \centering
    \includegraphics[width=0.8\linewidth]{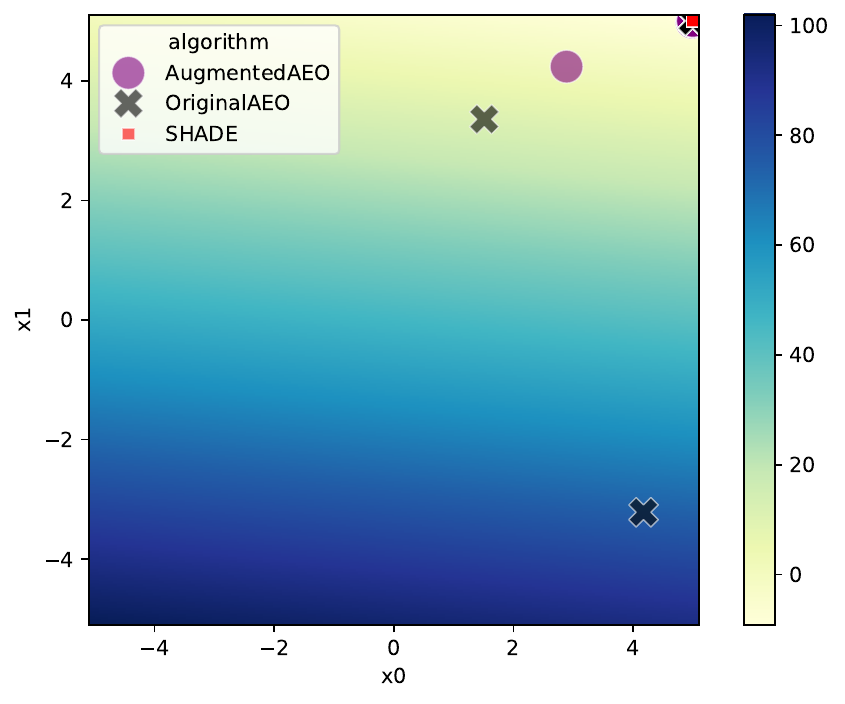}
    \caption{Visualization of the fifth 2$d$ BBOB problem (Linear Slope) and the candidate solutions explored by the OriginalAEO, AugmentedAEO and SHADE algorithms in the last iteration}
    \label{fig:2d_solution_plot_f5}
\end{figure}

\begin{figure}
    \centering
    \includegraphics[width=0.8\linewidth]{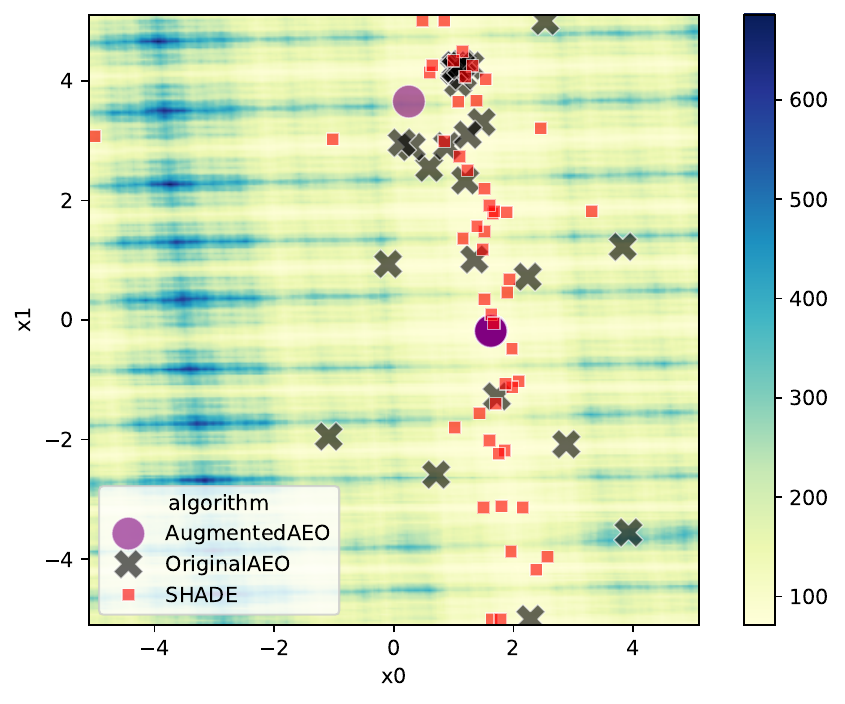}
    \caption{Visualization of the 16th 2$d$ BBOB problem (Weierstrass) and the candidate solutions explored by the OriginalAEO, AugmentedAEO and SHADE algorithms in the last iteration}
    \label{fig:2d_solution_plot_f16}
\end{figure}

To validate our representations, we visualize the solutions explored by the three algorithms in two dimensions. Figures~\ref{fig:2d_solution_plot_f5} and~\ref{fig:2d_solution_plot_f16} contain a visualization of the two previously investigated functions (Linear Slope and Weierstrass). The horizontal and vertical axes denote the two dimensions of the candidate solution, while the color reflects the objective function value of the solution (lighter is lower and better).
The location of the candidate solutions explored by the three algorithms in the last iteration of the search is denoted in the form of points on top of the color map. In Figure~\ref{fig:2d_solution_plot_f5}, most of the solutions are located in the upper right corner and are overlapping, meaning that the algorithms converge to the same solution, which we also saw in Figure~\ref{fig:single_traj_f5}. The AugmentedAEO algorithm has one candidate solution which is different from the majority of overlapping solutions located in the upper right corner, while the OriginalAEO algorithm has two such candidate solutions. This is also consistent with our findings in Figure~\ref{fig:single_traj_f5}, where these two algorithms had exactly the same number of candidate solutions located in different clusters.

Focusing on Figure~\ref{fig:2d_solution_plot_f16}, containing the same visualization for the Weierstrass problem, we can see that in this case, most of the candidate solutions are not overlapping. The exception is the AugmentedAEO algorithm, which has most of the candidate solutions overlapping around the point (1.8, 0) and a single differing solution around the point (0.5, 3.8). The other two algorithms have solutions in different area of the search space. This is also consistent with Figure~\ref{fig:single_traj_f16}.

\subsection{Analyzing the sensitivity of the optimization algorithm to its initialization}
In this subsection, we demonstrate the use of the proposed representations to evaluate the sensitivity of the optimization algorithm to its initialization, which we further refer to as stability.
To this end, we analyze the execution of a single algorithm across different runs (multiple executions with a different initial population). We do this by aggregating the cosine similarity of the proposed trajectory representation for trajectories obtained by a single algorithm executed on a single problem with different random seeds which impact the initialization of the initial population, or more precisely:
\begin{equation}
\begin{split}
    stability(a, p) = \frac{1}{|R| (|R| - 1)} \\
\sum_{\substack{r_k, r_l \in R \\ r_k \neq r_l}} sim(v_{a, p, r_k}, v_{a, p, r_l})
\end{split}
\label{eq:stability}
\end{equation}

where $a$ denotes an algorithm, $p$ denotes a problem, $R$ is the set of all random seeds used for different algorithm executions, $sim$ represents the cosine similarity metric, while $v_{a, p, r_{k}}$ is the vector representation of the trajectory of algorithm $a$ executed on problem $p$ using the random seed $r_k$.
The intuition behind this metric is that algorithms which have high similarity of trajectory representations when executed multiple times on the same problem are more stable than ones with lower similarities.

As an example, we demonstrate the behaviour of the AugmentedAEO and BaseDE algorithms on the first instance of the second BBOB 2$d$ problem class. According to our analysis, AugmentedAEO has a relatively unstable behaviour on this problem instance, with a stability score of 0.32. Conversely, the BaseDE algorithm has a stable behaviour with a stability score of 0.81.

\begin{figure}
\centering

\begin{subfigure}[b]{\linewidth}
\centering
\includegraphics[width=0.55\linewidth]{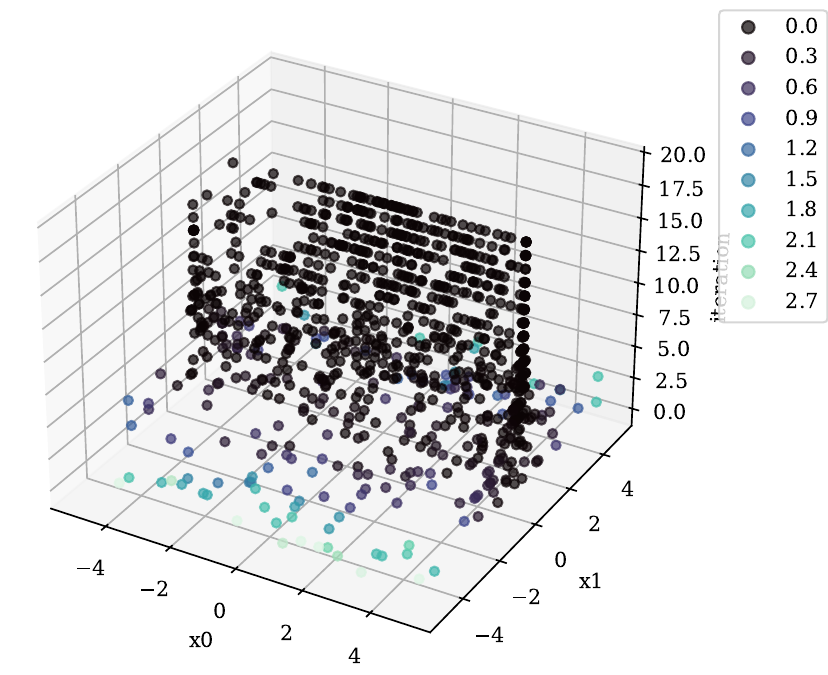}
\caption{BaseDE algorithm, first run}
\label{fig:baseDE_run_0}
\end{subfigure}

\begin{subfigure}[b]{\linewidth}
\centering
\includegraphics[width=0.55\linewidth]{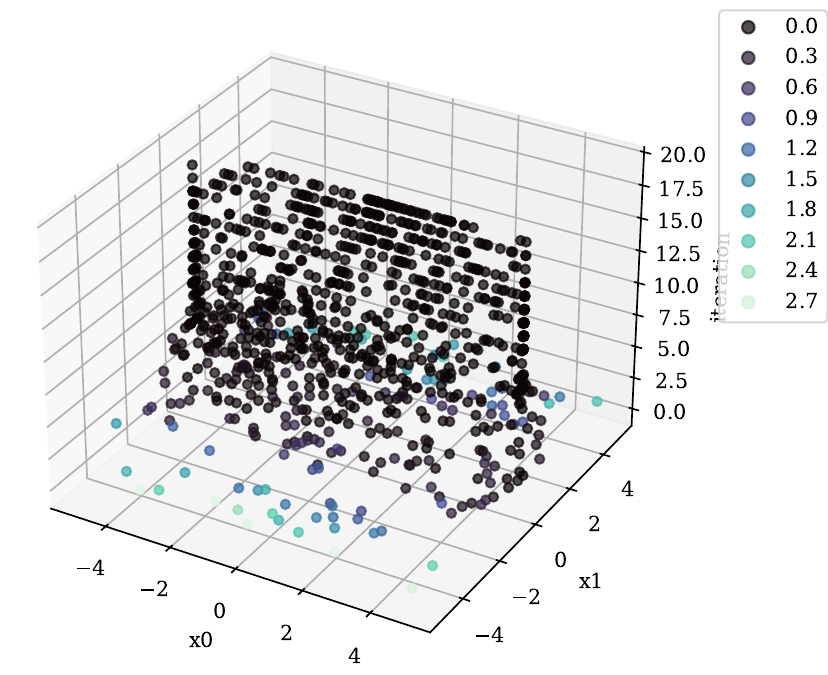}
\caption{BaseDE algorithm, second run}
\label{fig:baseDE_run_1}
\end{subfigure}

\begin{subfigure}[b]{\linewidth}
\centering
\includegraphics[width=0.55\linewidth]{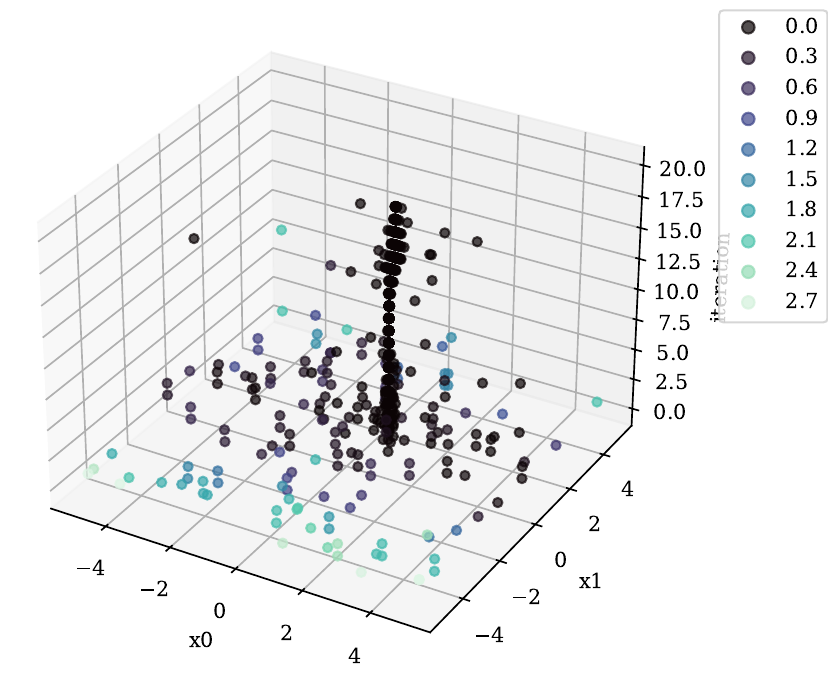}
\caption{AugmentedAEO algorithm, first run}
\label{fig:augmentedAEO_run_0}
\end{subfigure}

\begin{subfigure}[b]{\linewidth}
\centering
\includegraphics[width=0.55\linewidth]{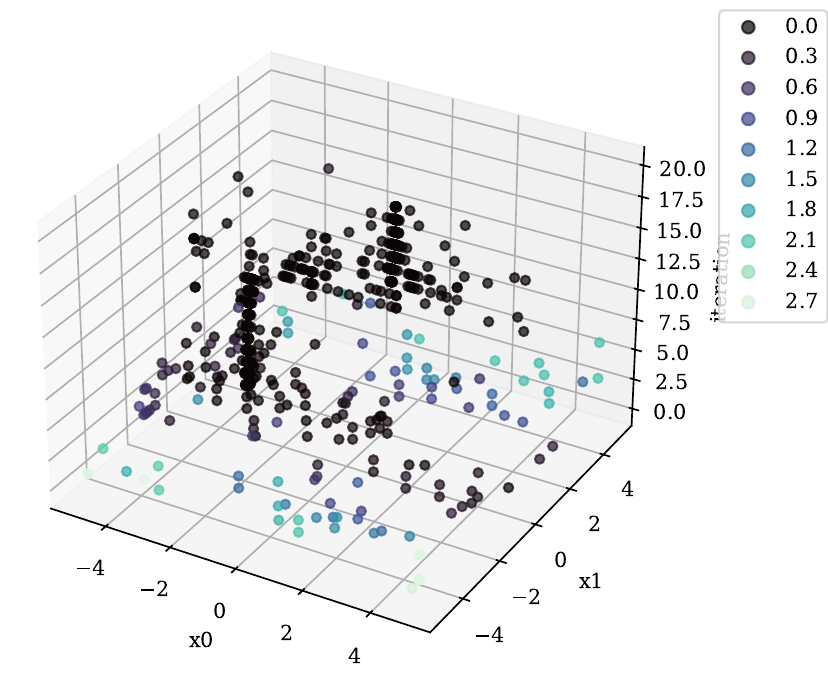}
\caption{AugmentedAEO algorithm, second run}
\label{fig:augmentedAEO_run_1}
\end{subfigure}

\caption{Trajectories of the BaseDE and AugmentedAEO algorithms executed three times on the first instance of the second BBOB 2$d$ problem class}
\label{fig:example_trajectories_scatterplot}
\end{figure}

Figure~\ref{fig:example_trajectories_scatterplot} shows the trajectories of these two algorithms produced in two different executions on the first instance of the second BBOB 2$d$ problem class. The visualization is generated following~\cite{tea_visualizing}, with the exception that we do not perform any dimensionality reduction, since we have a 2$d$ problem.
Each subplot contains a single trajectory. The axes on the bottom, with range [-5,5], represent the values of the candidate solutions in the two dimensions. The vertical axis, with range [0,20], represents the iteration number, while the color indicates the objective function value of the candidate solution. Comparing figures ~\ref{fig:baseDE_run_0} and~\ref{fig:baseDE_run_1}, plotting two runs of the BaseDE algorithm obtained with two different random seeds, we can see that the distribution of the candidate solutions is similar. In both runs, the algorithm does not converge to a single solution, but finds several solutions of similar quality. The locations of the explored candidate solutions are similar in both runs, regardless of the starting population.
On the contrary, comparing the two executions of the AugmentedAEO algorithm in figures~\ref{fig:augmentedAEO_run_0} and~\ref{fig:augmentedAEO_run_1}, we can see a differing search pattern. In Figure~\ref{fig:augmentedAEO_run_0}, after the fifth iteration, the AugmentedAEO algorithm is mainly doing exploitation around the point (1.2, 0.4). On the other hand, in Figure~\ref{fig:augmentedAEO_run_1}, from iteration 3 to iteration 11, most of the candidate solutions are located around the point (-3.2, 0.4). In iterations 12 to 16, the algorithm jumps to another area of the search space, located around the point (1.3, 0.4).
This means that the behaviour of the AugmentedAEO algorithm is more affected by the initialization of the first population, i.e., it is less stable than the BaseDE algorithm.

\begin{figure}
    \centering
    \includegraphics[width=0.8\linewidth]{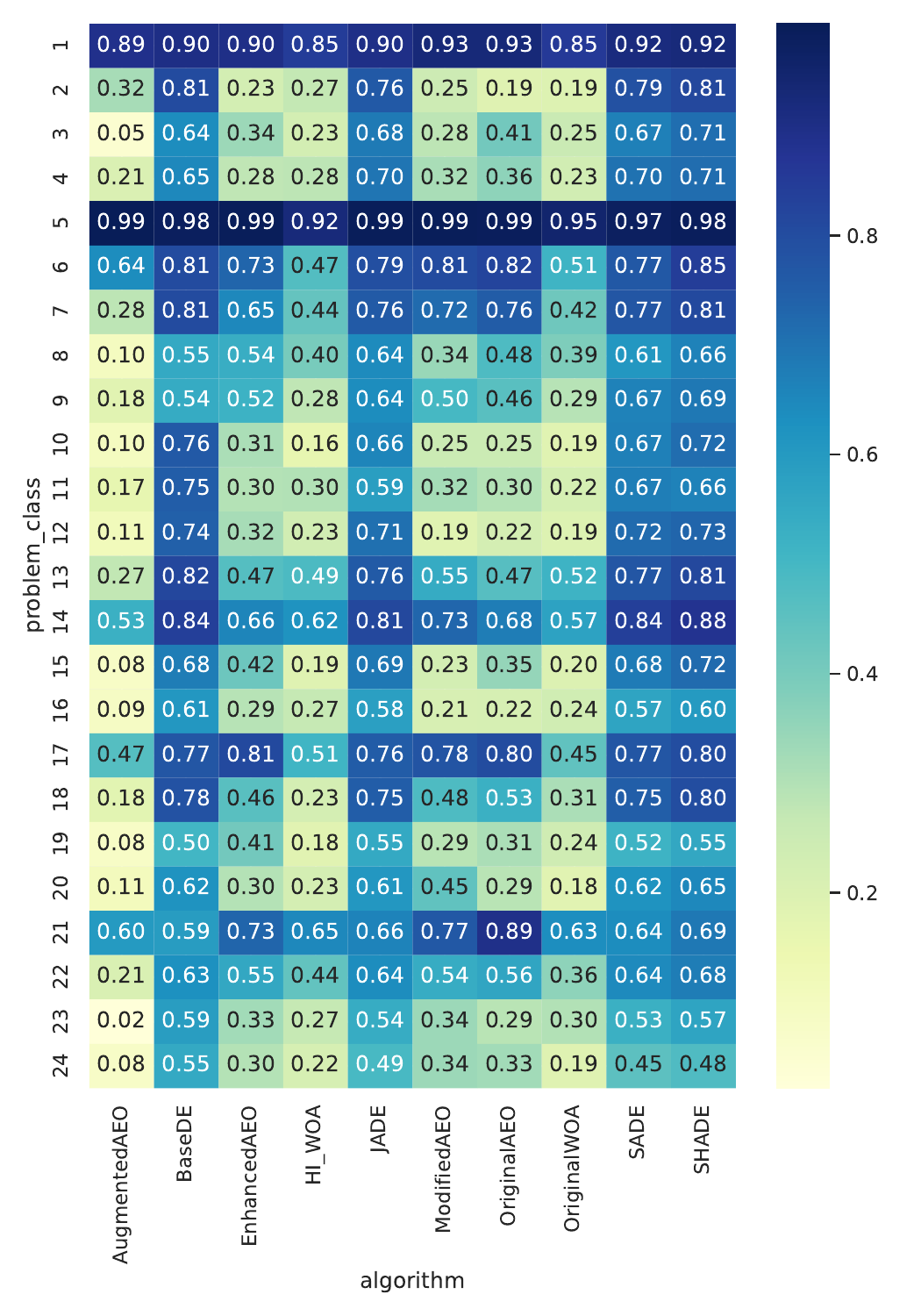}
    \caption{Algorithm stability per problem, 2$d$}
    \label{fig:algorithm_stability_per_problem_2d}
\end{figure}

\begin{figure}
    \centering
    \includegraphics[width=0.8\linewidth]{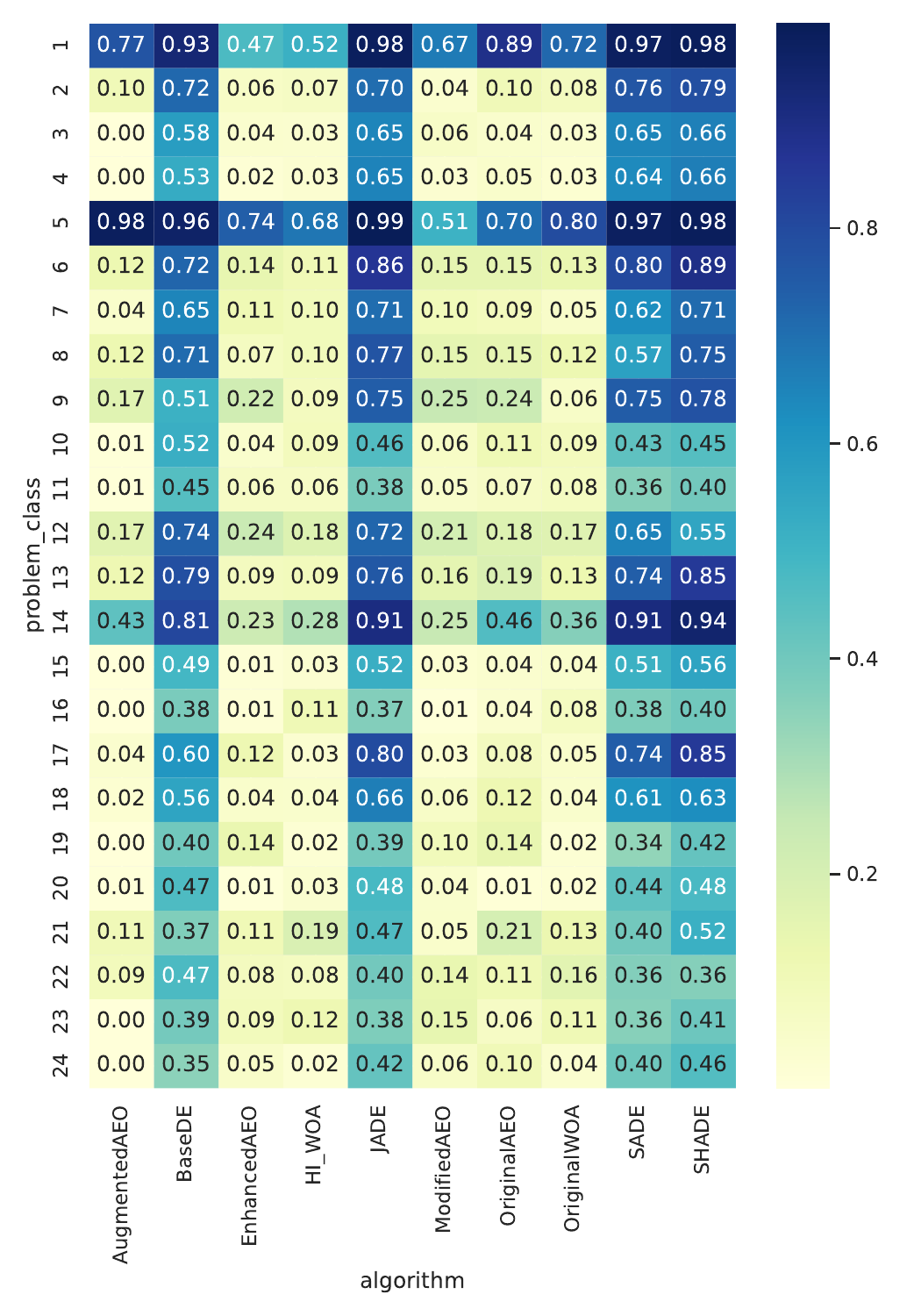}
      \caption{Stability of the investigated algorithms on the 5$d$ BBOB problems, averaged across problem instances}
   \label{fig:algorithm_stability_per_problem_5d}
\end{figure}

Figures~\ref{fig:algorithm_stability_per_problem_2d} and ~\ref{fig:algorithm_stability_per_problem_5d} show the mean algorithm stability scores (aggregated across all instances of a problem class) for each algorithm on each of the BBOB problems in dimension 2, and 5, respectively.
We can see that there are problems where all of the algorithms have a relatively stable behaviour, and problems where some algorithms experience lower stability. From Figure~\ref{fig:algorithm_stability_per_problem_2d} we can see that in dimension 2, most algorithms are relatively stable on problems 1, 5, 6, 14, 17 and 21. This is especially pronounced for problems 1 and 5, where all algorithms have stability scores above 0.85. Considering the simplistic nature of these two functions, it does make sense that most algorithms achieve stable behavior and smoothly find the optimum across all runs. From Figure~\ref{fig:algorithm_stability_per_problem_5d} we observe that in dimension 5, the only two problems where all algorithms are relatively stable are 1 and 5. In this dimension the DE-based algorithms (BaseDE, JADE, SADE and SHADE) have much higher stability scores than the other algorithms, which are more affected by the initialization. The same pattern is also present for 10$d$ problems, although we omit the figure due to space limitations. This also makes sense, since the problem dimension increases the complexity and size of the search space, likely resulting in a larger number of clusters.

\subsection{Analyzing the similarity of different optimization algorithms}
When calculating the similarity of different optimization algorithms, we take the mean similarity of the algorithm trajectories produced by a pair algorithms on the same problem using the same random seed (starting with the same initial population). In particular, the similarity of a pair of algorithms $a_m$ and $a_n$ is calculated as:
\begin{equation}
\begin{split}
     similarity(a_m,a_n) = \frac{1}{|R||P|} \\ \sum_{r \in R} \sum_{p \in P} sim\big(v_{a_m, p, r}, v_{a_n, p, r}\big) 
\end{split}
   \label{similarity}
\end{equation}

where $R$ is the set of all random seeds used for algorithm execution, $P$ is the set of all problems, $sim$ is the cosine similarity metric, while $v_{a_m, p, r}$ is the vector representation of the trajectory of algorithm $a_m$ executed on problem $p$ using the random seed $r$. Note that here we are comparing trajectories of different algorithms initialized with the same population (same random seed $r$), whereas in Equation~\ref{eq:stability} we are comparing trajectories of the same algorithm initialized with different populations (different random seeds). We do this since when we are measuring stability, we are interested in the behaviour of a single algorithm, accross different initializations. On the other hand, when measuring the similarity of two algorithms, we want to eliminate the effect of the random initialization on the produced trajectories, so we therefore compare the algorithms starting from the same initial population, i.e, same random seed.

\begin{figure}
  \centering
  \begin{subfigure}[b]{0.85\linewidth}
  \centering
       \includegraphics[width=\linewidth]{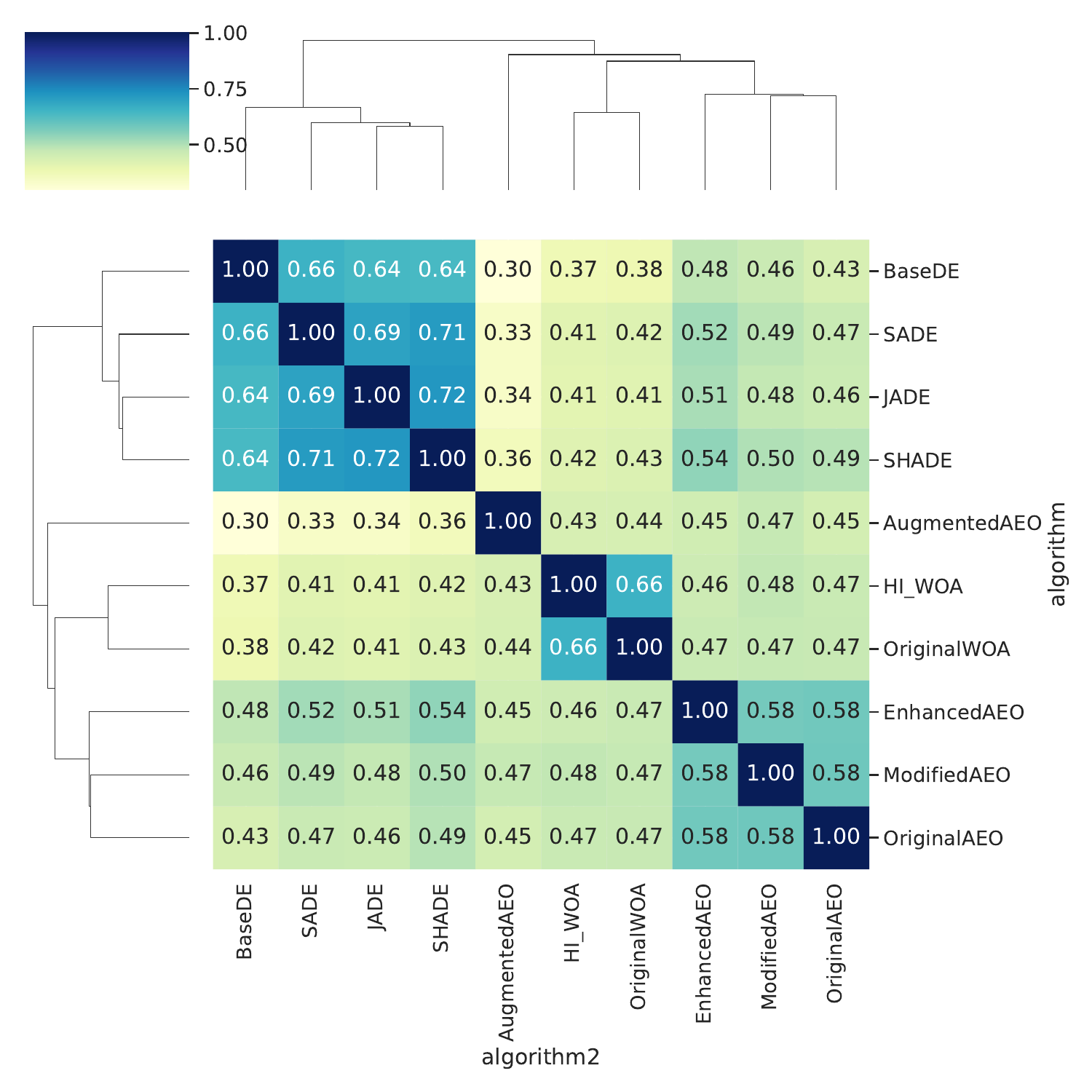}
       \label{fig:algorithm_similarity_2d}
   \end{subfigure}
    \hfill
   \begin{subfigure}[b]{0.85\linewidth}
       \centering
       \includegraphics[width=\linewidth]{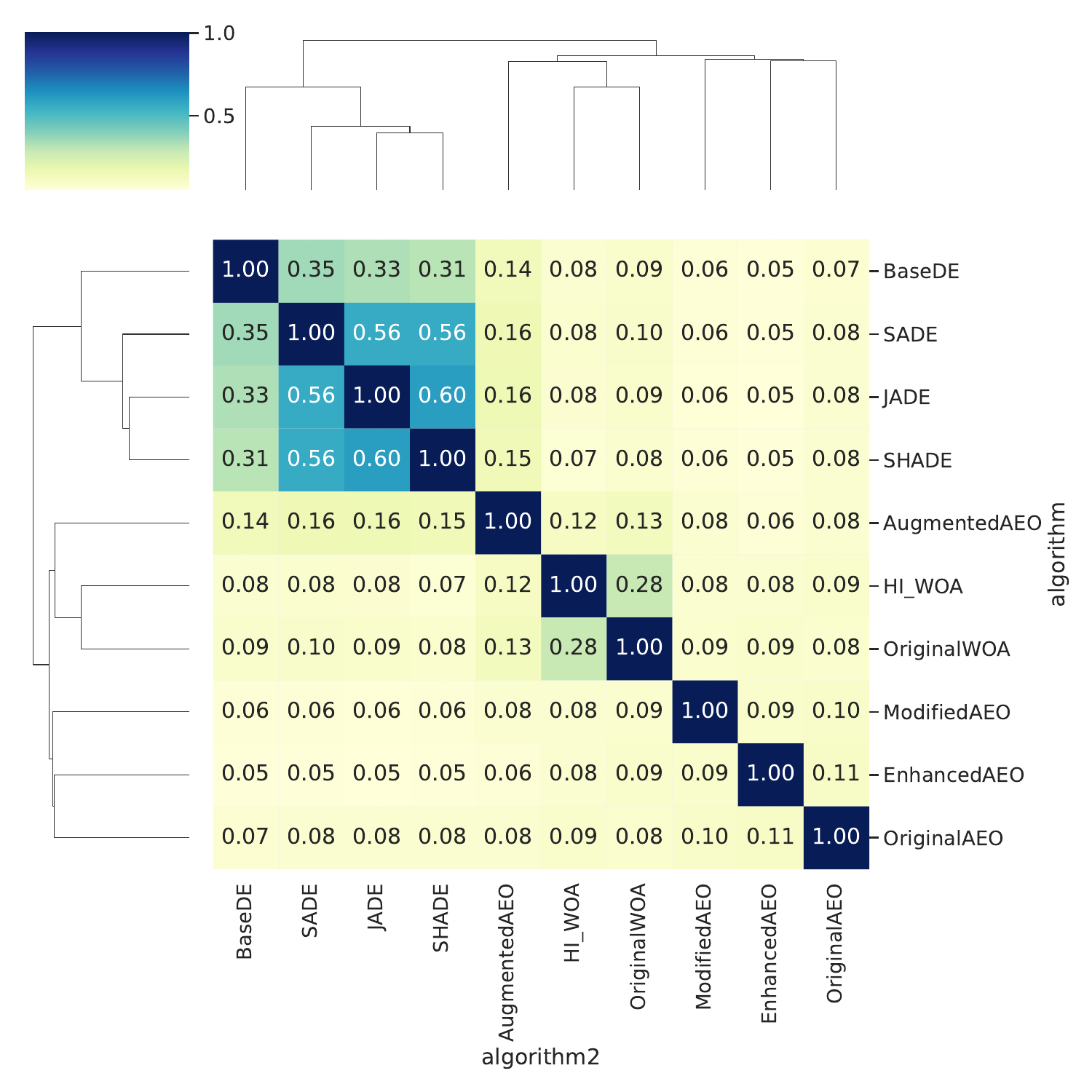}
        \label{fig:algorithm_similarity_10d}
    \end{subfigure}
    \caption{Algorithm similarity across problem dimensions for 2$d$ (top), and 10$d$ (bottom) problems.}
    \label{fig:algorithm_similarity}
\end{figure}

Figure~\ref{fig:algorithm_similarity} presents the mean algorithm similarity on 2$d$, and 10$d$ problems. In the first subplot, containing the similarity on 2$d$ problems, we can see that the DE-based algorithms (BaseDE, SADE, JADE and SHADE) are clustered together and have similarity scores in the range 0.64 -- 0.72. The HI\_WOA and OriginalWOA algorithms also have a quite high similarity score of 0.66. Three of the AEO-based algorithms (EnhancedAEO, ModifiedAEO and OriginalAEO) are clustered together and have a similarity score of 0.58. Interestingly, the AugmentedAEO algorithm bears a somewhat lower similarity to the other AEO variants. We can also observe that the similarities of variants of the same algorithm are generally higher than the similarities of variants of a different algorithm, which indicates that our proposed representations are able to capture algorithm similarities in a meaningful way.
Focusing on the results for 10$d$ problems, we can see that the DE variants and the WOA variants are still clustered together, although with somewhat lower similarity scores. On the other hand, the AEO variants have relatively low similarity scores with any other algorithms. This is to some degree consistent with our analysis of their stability, where these algorithms become unstable in higher problem dimensions.
An additional analysis of the execution time of each step of the methodology can be found in our github repository.

\section{Conclusion}
\label{sec:conclusion}
 
This paper introduced ClustOpt, a novel vectorized representation and visualization methodology for analyzing the search process of optimization algorithms by clustering solution candidates and tracking their cluster transitions across iterations. This approach facilitated the definition of two key metrics: stability, which quantifies the consistency of an algorithm's trajectories across different initial populations, and similarity, which measures the resemblance of search trajectories between different algorithms. 
By offering a means to track, quantify, and visualize the dynamic behaviors of optimization algorithms, our framework provides valuable insights into their exploratory and exploitative characteristics, sensitivity to initialization, and comparative behaviors, thereby enabling a deeper understanding and refinement of optimization strategies. 
For future work, we are planning to perform a comprehensive analysis of the entire MEALPY Python library, to empirically find variants of the same algorithm leading to the same outcomes, so it can be a step further in addressing the call for action related to the methaphor-based metaheuristics.
Finally, we would like to point out that this is an introductory study to the proposed methodology and a more detailed analysis of the impact of the clustering algorithm, its parameters, and similarity metric would be useful to further understand its behaviour, outcomes and limitations.
\section*{Acknowledgments}
We acknowledge the support of the Slovenian Research and Innovation Agency through program grant No.P2-0098, young researcher grants No.PR-12393 to GC and No.PR-11263 to GP, and project grants No.J2-4460 and No.GC-0001. This work is also funded by the European Union under Grant Agreement No.101187010 (HE ERA Chair AutoLearn-SI) and the EU Horizon Europe program (grant No.101077049, CONDUCTOR).

\bibliographystyle{IEEEtran}
\bibliography{references}

\end{document}